\documentclass{article}
\usepackage{spconf,amsmath,graphicx,url}
\usepackage{caption}
\usepackage{multirow}
\usepackage{float}
\usepackage{booktabs}
\usepackage{amsmath}
\usepackage{lipsum}

\usepackage{cite}
\usepackage{microtype}

\def\ie{{i.e.}}

\def\etal{{\emph{et al.~}}}

\title{CONTOUR-WEIGHTED LOSS FOR CLASS-IMBALANCED IMAGE SEGMENTATION}
\name{Zhengyong Huang\textsuperscript{1,2}, Yao Sui\textsuperscript{1,2,} \thanks{This work was supported by the Faculty Development Award from Peking University under Award No. 71013Y2268.}\sthanks{Corresponding author: suiyao@hsc.pku.edu.cn (Y. Sui).}}
\address{\textsuperscript{1}National Institute of Health Data Science, Peking University, Beijing, China\\
\textsuperscript{2}Institute of Medical Technology, Peking University, Beijing, China
}
\begin{document}
%

\setlength{\lineskiplimit}{0pt}
\setlength{\lineskip}{0pt}
\setlength{\abovedisplayskip}{3pt}   
\setlength{\belowdisplayskip}{3pt}
\setlength{\abovedisplayshortskip}{6pt}
\setlength{\belowdisplayshortskip}{6pt}

\maketitle
\begin{abstract}
Image segmentation is critically important in almost all medical image analysis for automatic interpretations and processing. However, it is often challenging to perform image segmentation due to data imbalance between intra- and inter-class, resulting in over- or under-segmentation. Consequently, we proposed a new methodology to address the above issue, with a compact yet effective contour-weighted loss function. Our new loss function incorporates a contour-weighted cross-entropy loss and separable dice loss. The former loss extracts the contour of target regions via morphological erosion and generates a weight map for the cross-entropy criterion, whereas the latter divides the target regions into contour and non-contour components through the extracted contour map, calculates dice loss separately, and combines them to update the network. We carried out abdominal organ segmentation and brain tumor segmentation on two public datasets to assess our approach. Experimental results demonstrated that our approach offered superior segmentation, as compared to several state-of-the-art methods, while in parallel improving the robustness of those popular state-of-the-art deep models through our new loss function.
The code is available at \url{https://github.com/huangzyong/Contour-weighted-Loss-Seg}.

\end{abstract}
\begin{keywords}
image segmentation, medical image computing, contour weighting map, deep learning, loss function
\end{keywords}
\section{INTRODUCTION}
\label{sec:intro}

Image segmentation plays a critically important role in medical image analysis. It is in general the first step for quantitative analysis of anatomical structures \cite{hatamizadeh2022unetr}. Automatic labeling organs and structures of interest are often necessary to perform tasks such as clinical diagnosis, radiomics, and personalized medicine \cite{huang2022isa}. Therefore, the performance of image segmentation significantly affects the quality of medical image analysis. Recent years have witnessed considerable improvement in image segmentation \cite{cao2022swin}. There are, unfortunately, still gaps required to fill between techniques and application scenarios \cite{chen2022recent}. One of the most prominent issues is the data imbalance between intra- and inter-class \cite{wang2023dhc}, which commonly raises over- and/or under-segmentation errors \cite{zhang2021rethinking}.

Data imbalance inevitably results in false positives or false negatives. Therefore, it is desired to find a trade-off in practical application scenarios. To this end, various methods have been developed for improving image segmentation, including threshold-based strategies \cite{guo2020learn}, region-based approaches \cite{meng2021graph}, morphological methods \cite{wang2020boundary}, and deep learning \cite{liu2023clip, hu2023label}. Among those methods, an intuitive approach to deal with data imbalance is to design a sophisticated network architecture. U-Net \cite{ronneberger2015u} is one of the most effective frameworks in this category, which comprises an encoder and a decoder network. Although U-Net-like models currently stand as an important backbone for medical image segmentation and serve as a foundational inspiration for the development of several variant networks \cite{cao2022swin, wang2023densely}, it is too challenging and complicated to equip the architecture with particular structures that handle data imbalance issues, leading to complex models difficult to train and time-consuming.

Leveraging improved loss functions has recently emerged as an effective strategy that allows for alleviating data imbalance issues efficiently. The latest results demonstrated that using improved loss functions in image segmentation enables lightweight model design, allows for robustness promotion, improves model generalization, and in turn facilitates practical applications \cite{zhao2022act}. Current methods based on improved loss functions can mainly be classified into three categories: distribution-based, region-based, and compound loss functions \cite{zhu2022compound}. 
Distribution-based loss functions, such as cross-entropy loss \cite{wang2019symmetric}, minimize the differences between two probability distributions. Although these methods in this category facilitate controlling false positives and false negatives, they are not sufficiently stable when dealing with images suffering from a severe imbalance between classes. Ronnebrger \etal \cite{ronneberger2015u} proposed a distance map-weighted cross-entropy loss (DWCE) to handle intra-class imbalance and achieved promising results. They assigned larger weights to those pixels near the boundary than pixels far away from the boundary for constructing a weighted cross-entropy loss. Region-based loss functions, such as dice loss \cite{milletari2016v}, aim to maximize the overlap between the predicted and ground truth images. These functions pay more attention to the foreground and therefore mitigate inter-class imbalance issues. Sudre \etal \cite{2017Generalised} proposed a generalized dice loss (GDL) to address intra-class imbalances. They generated a weighting map according to the volume of each target region. Compound loss functions, such as combo loss \cite{taghanaki2019combo,zhao2022act} that integrates cross entropy and dice loss (CEDL), combine the advantages of different types of loss functions and thus are effective for both inter-class and intra-class imbalance.

Inspired by the previous success, we developed a new methodology to construct a new loss function to improve the segmentation performance for medical images. We started from the analysis of two difficulties that commonly occur in medical image segmentation. One is that the target foreground is too small, where the distribution of the target foreground and background is unbalanced. Retinal vascular segmentation is such a case \cite{fu2023robust}. The other is that there are multiple segmentation targets with variable sizes, so the distribution between each category is unbalanced, such as risk organ segmentation \cite{zhang2023continual} and brain tumor segmentation \cite{do20233d}. As a result, we considered promoting the accuracy in the locations of the boundary contours, in order to deal with the above difficulties, and consequently to mitigate the intra- and inter-class imbalance, leading to robust segmentation. Therefore, we proposed a new compound loss function with a contour-weighted strategy that utilizes cross-entropy and dice loss (CWCD). We generated contour-weighted maps to strengthen the deep neural networks with a focus on segmentation boundaries, resulting in improved segmentation accuracy. We carried out abdominal organ segmentation and brain tumor segmentation on two public datasets to assess our approach. Experimental results demonstrated that our approach offered superior segmentation, as compared to several state-of-the-art methods, while in parallel improving the robustness of those popular state-of-the-art deep models through our new loss function.

Our main contributions are summarized as follows:
\begin{itemize}
\item[$\bullet$] We established the concept of contour weighting for image segmentation and found that it is effective to solve the intra- and inter-classes data imbalance issue.
\item[$\bullet$] We proposed a new compound loss function that combines contour-weighted cross-entropy and separate dice loss for medical image segmentation.
\item[$\bullet$] We demonstrated the effectiveness of our method in multiple segmentation tasks on two public datasets.
\end{itemize}

\begin{figure}[t]
    \centering
    \setlength{\belowcaptionskip}{-0.3cm}
    \includegraphics[width=3.4in]{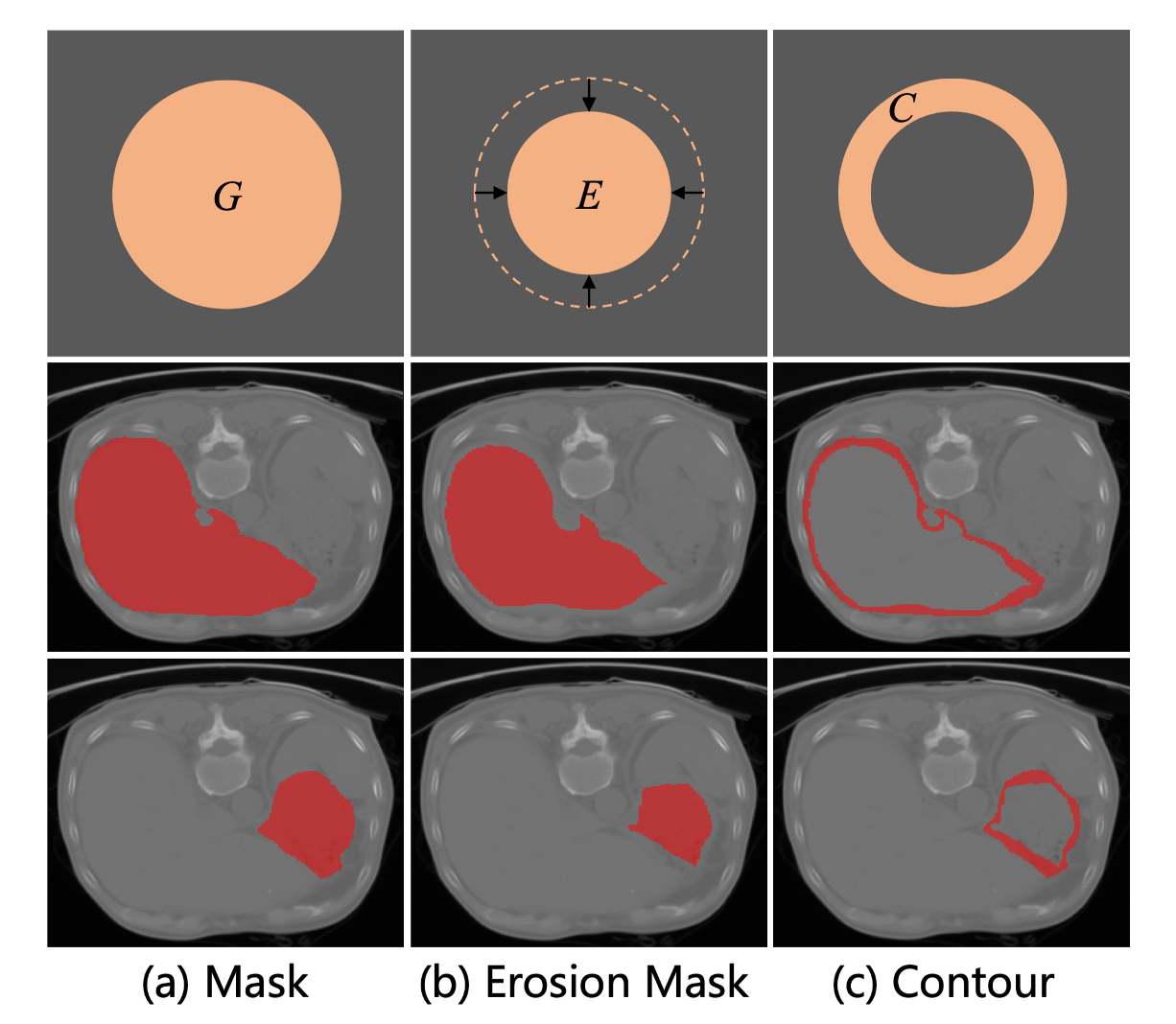}
    
    \caption{Illustration of our proposed contour-weighted map. We calculate the contours $C$ from the difference between the mask $G$ and its morphological erosion correspondence. The obtained contours $C = G - E$ are used to weight the cross-entropy loss and compute the separable dice loss. The second and third rows show the contour calculation process on two representative images, respectively.}
    
    \label{contour}
\end{figure}
\section{METHODS}
\label{sec:pagestyle}
We develop a novel compound loss function to address the challenges of intra- and inter-class imbalance, with the aim to improve the segmentation performance for medical images. Our loss function comprises two components: a separable dice loss and a contour-weighted cross-entropy loss.
\subsection{Contour Extraction}
\label{Contour Extraction}
We construct the contour weighting map on the label image for our loss function. We calculate the contours of the segmentation targets by using morphological operations. Morphological erosion is able to eliminate the boundary points of a connected region so that the boundary shrinks inward. Fig.\ref{contour} illustrates the calculation strategy for the contour of the segmentation target. Specifically, we use morphological erosion to push the target boundary to shrink inward and the target volume becomes smaller. Then, the eroded target is subtracted from the original target to obtain the contour. This process is described as
\begin{equation}
\begin{aligned}
\label{contour extraction}
    C = G - (G \ominus k) 
\end{aligned}
\end{equation}
where $G$ denotes the segmentation target, $\ominus$ denotes the morphological erosion operation, and $k$ denotes the erosion kernel of size $3 \times 3 \times 3$ pixels, and the number of iterations is 6.

\vspace{-0.2cm}
\subsection{Separable Dice Loss}
\label{Separable Dice Loss}
There are commonly multiple target regions to be segmented in medical image segmentation tasks. It is challenging for learning models to accurately capture all target regions in the presence of data imbalance between those target regions. The learning may converge to a local minima of the loss function. So, the trained network prefers picking a single or partial target region and may overlook the remaining regions.

Dice similarity coefficient (DSC), ranging from 0 to 1, is often used to evaluate segmentation results. The dice similarity coefficient of a volumetric image with $N$ binary voxel values is found by
\begin{equation}\label{dice loss}
    \mathcal{DSC} = \frac{2 \sum_{i=1}^{N} p_{i} g_{i}} {\sum_{i=1}^{N} p_{i}^2+\sum_{i=1}^{N} g_{i}^2 + \epsilon },
\end{equation}
where $p_i$ and $g_i$ denote the $i$-th voxel from the predicted and ground truth segmentation, respectively, and $\epsilon$ is a small constant to avoid division by 0. A variant is designed as follows to take care of those segmentation tasks with a small number of target regions \cite{milletari2016v}.
%
\begin{equation}\label{dice loss}
    \mathcal{L}_{Dice} = 1 - 2 \frac{\sum_{i=1}^{N} p_{i} g_{i}} {\sum_{i=1}^{N} p_{i}^2+\sum_{i=1}^{N} g_{i}^2 + \epsilon }.
\end{equation}

However, as the number of target regions increases, the imbalance between those regions is more prominent, and unfortunately, the dice loss becomes less effective. Consequently, we design a separable dice loss function to mitigate the imbalance issue by generating a weighting map for the segmentation according to the boundaries of target regions. Considering that the segmentation errors are attributed to two factors: over-segmentation and under-segmentation, and in either case, the error regions are concentrated at the edge positions. In our separable dice loss (SDL) function, we separate the target regions into contour and non-contour (erosion mask) components (Fig. \ref{contour}) and then calculate the dice loss of the two components separately, to increase the weight of the segmentation boundary.
\begin{equation}\label{contour dice loss}
    \mathcal{L}_{c} = 1- \frac {1}{M} \sum_{j=1}^{M} \frac{2 \sum_{i=1}^{N_1} {p_{i,j}^1}^2 {g_{i,j}^1}^2}{\sum_{i=1}^{N_1} {p_{i,j}^1}^2+\sum_{i=1}^{N_1} {g_{i,j}^1}^2 + \epsilon },
\end{equation}
\begin{equation}\label{noncontour dice loss}
    \mathcal{L}_{noc} = 1- \frac {1}{M} \sum_{j=1}^{M} \frac{2 \sum_{i=1}^{N_2} {p_{i,j}^2}^2 {g_{i,j}^2}^2}{\sum_{i=1}^{N_2} {p_{i,j}^2}^2+\sum_{i=1}^{N_2} {g_{i,j}^2}^2 + \epsilon },
\end{equation}
\begin{equation}\label{sDice loss}
    \mathcal{L}_{SDL} = \lambda \mathcal{L}_{c} + (1-\lambda)\mathcal{L}_{noc},
\end{equation}
where $M$ denotes the number of target regions, $g^1$ and $g^2$ denote the separated contour and non-contour, respectively, $g^1 \cup g^2\in Mask$, $N_1 + N_2 = N$. Similarly, $p^1$ and $p^2$ denote the predicted contour and non-contour, respectively. $\lambda $ is a weight parameter that controls the importance of the contour component, and $\lambda$ is empirically set to 0.5 in this paper.

The value of $\mathcal{L}_{noc}$ decreases faster than that of $\mathcal{L}_{c}$ during the network training. Therefore, the segmentation errors are dominated by the target edges. $\mathcal{L}_{c}$ is thus more important when updating the gradient for training. To this end, our separable dice loss is able to effectively handle inter-class imbalances because it pays more attention on the overlap between predicted and ground truth masks for each individual class. 

\vspace{-0.3cm}
\subsection{Contour-Weighted Cross Entropy Loss}
\label{Contour Weighted Cross Entropy Loss}
Cross-entropy loss \cite{mao2023cross} is widely used in deep learning, in particular intensively leveraged in the deep neural networks for image segmentation. Its effectiveness in reducing the distance between the predicted and actual probability distributions makes it a preferred choice for optimizing neural networks. However, in the presence of data imbalance, cross-entropy loss function may lead to biased training results. The binary cross-entropy loss is defined as
\begin{equation}\label{ce loss}
    \mathcal{L}_{CE}^b = -\sum_{i=1}^{N}(g_{i}\log p_{i} + (1-g_i)\log (1-p_i)).
\end{equation}
We rewrite Eq.(7) as
\begin{equation}\label{ce loss}
    \mathcal{L}_{CE}^b = -\sum_{i=1}^{N}g_{i}\log p_{i}.
\end{equation}
When there are multiple categories of target regions to be segmented, the cross-entropy loss is found by
\begin{equation}\label{ce loss}
    \mathcal{L}_{CE}^m = - \frac {1}{M} \sum_{j=1}^{M} \sum_{i=1}^{N}g_{i}\log p_{i}.
\end{equation}
To mitigate the intra-class imbalance, we design a cross-entropy loss function based on a contour weighting map, defined as:
\begin{equation}\label{ce loss}
    \mathcal{L}_{CE}^w = - \frac {1}{M} \sum_{j=1}^{M} \sum_{i=1}^{N} {w_{c}}_{(i)}  g_{i}\log p_{i},
\end{equation}
where ${w_{c}}$ denotes the contour weight map we extracted using the method described in \ref{Contour Extraction}. The contour-weighted cross-entropy assigns large weights to those pixels belonging to the contours, so emphasizes the importance of the boundaries of target regions.

We obtain the compound loss function from
\begin{equation}\label{ce loss}
    \mathcal{L} = \mathcal{L}_{SDL} + \mathcal{L}_{CE}^w
\end{equation}

The integration of these two loss components deals with both intra- and inter-class imbalances. The compound loss function enhances the robustness and accuracy of segmentation with multiple categories of target regions.

\subsection{Implementation Details and Datasets}
We assessed our proposed approach on two benchmark datasets to perform brain tumor segmentation (BraTS) \cite{baid2021rsna} and abdominal organ segmentation (AMOS) \cite{ji2022amos}. The BraTS dataset has 1251 multi-modal MRI data (Flair, T1w, T2w, T1ce) with 3 labels of necrotic tumor core (NTC), peritumoral edema (ED), and enhancing tumor (ET) and is divided into 800, 200 and 251 for training, validation, and testing. However, we pay more attention to the segmentation results of the whole tumor (WT, a union of NTC, ED, and ET), the enhanced tumor (ET), and the tumor core (TC, a union of ET and ED). The AMOS dataset has 221 CT scans containing 15 labels, including spleen (Sp), right kidney (RK), left kidney (LK), gallbladder (Ga), esophagus (Es), liver (Li), stomach (St), aorta (Ao), postcava (Po), pancreas (Pa), right adrenal gland (RAG), left adrenal gland (LAG), duodenum (Du), bladder (Bl) and prostate/uterus (P/U). All data volumes were resampled into isotropic voxel spacing of $1.0 \emph{mm} \times 1.0 \emph{mm} \times 1.0 \emph{mm}$. The 221 scans on the AMOS dataset are divided into 180, 20, and 21 scans for training, validation, and testing. In the training process, the BraTS dataset was cropped to a size of $128\times160\times160$ voxels, and the AMOS dataset was cropped to a size of $d\times256\times256$ voxels and $d$ is the number of the axial slices. The voxel intensities are pre-processed with Z-score normalization.
\begin{equation}
    \hat{y}_i = \frac{y_i - \mu}{\sigma}
\end{equation}
where $\mu$ and $\sigma$ denote the mean and standard deviation of the image $y$, respectively.

We employed U-Net as a backbone network and substituted the loss functions, following the protocol in \cite{wang2021head}, to assess our proposed loss function. The U-Net network we utilized comprised four down/up sampling layers. The first convolutional layer had 16 kernels, which we doubled/halved when the feature map resolution was halved/doubled in the following layers. ReLU activation functions were used in the intermediate convolutional layers followed by BatchNormalization. The output of the network had the same resolution as the input image.

\begin{table}[h]
    \scriptsize
    \centering
    \caption{Segmentation results based on U-Net using loss functions of GDL, DWCE, CEDL, and our proposed CWCD on the two datasets, respectively, in terms of DSC. The best results are highlighted in bold font.}
    \label{Tab 1}
    \renewcommand\arraystretch{1.05}
    \begin{tabular}{cccccc}
        \toprule
        \specialrule{0em}{0.2pt}{0.2pt}
        \cline{1-6}
        Dataset & Organ & GDL\cite{2017Generalised} & DWCE\cite{ronneberger2015u} & CEDL\cite{taghanaki2019combo} & CWCD\\ 
        \cline{1-6}
        	& Sp &0.8965	&0.8351	&0.8831	&0.9016  \\
                & RK &0.9241	&0.8661	&0.8961	&0.9242  \\
                & LK &0.9112	&0.8847 &0.8875 &0.9151  \\
                & Ga &0.5527	&0.6153	&0.6416	&0.6444  \\ 
                & Es &0.4794	&0.6634	&0.5170	&0.6462  \\ 
                & Li &0.9479	&0.8995	&0.9420	&0.9393  \\
                & St &0.7451	&0.6821	&0.7715	&0.8117  \\
        AMOS    & Ao &0.9023	&0.8926	&0.9051	&0.9124  \\
                & Po &0.8306	&0.8034	&0.8288	&0.8368  \\
                & Pa &0.6246	&0.6152	&0.5974	&0.6789  \\
                & RAG&0.5395	&0.6087	&0.4397	&0.6116  \\
                & LAG&0.3619	&0.6208	&0.3549	&0.4869  \\
                & Du &0.6156	&0.5911	&0.5942	&0.6408  \\
                & Bl &0.6879	&0.6381	&0.7120	&0.6810  \\
                & P/U &0.5501	&0.2671 &0.6842 &0.6149  \\
                & Avg &0.7046	&0.6989	&0.7103	&\textbf{0.7497} \\

        \cline{1-6}
        \rule{0pt}{8pt} 
                &WT	&0.8064	&0.8256	&0.8411	&0.8493  \\
        BraTS   &TC	&0.8292	&0.8081	&0.7868	&0.8182  \\
                &ET	&0.7855	&0.7572 &0.7447	&0.7622  \\
                &Avg&0.8070	&0.7970	&0.7909	&\textbf{0.8099}  \\     
        \cline{1-6}
        \specialrule{0em}{0.3pt}{0.3pt}
        \bottomrule
    \end{tabular}
\end{table}

We trained the network on the two datasets independently. We created the training dataset on AMOS by cropping the images in the axial direction with a size of $64 \times 256 \times 256$ voxels and an overlap of 16 slices between two consecutive volumetric crops. We incorporated data enhancement strategies in the training on both datasets, including random rotation (angle range [0, 15] degrees with probability of 0.2) and flips (horizontal or vertical flip, with probability of 0.2). We carried out 50 epochs for training using an Adam optimizer with an initial learning rate of 3e-4, and we halved the learning rate at the $20^{th}$ and $40^{th}$ epochs. The model with the best performance in terms of DSC on the validation set was selected for testing.
\begin{figure}[htbp]
    \centering
    \includegraphics[width=3.3in]{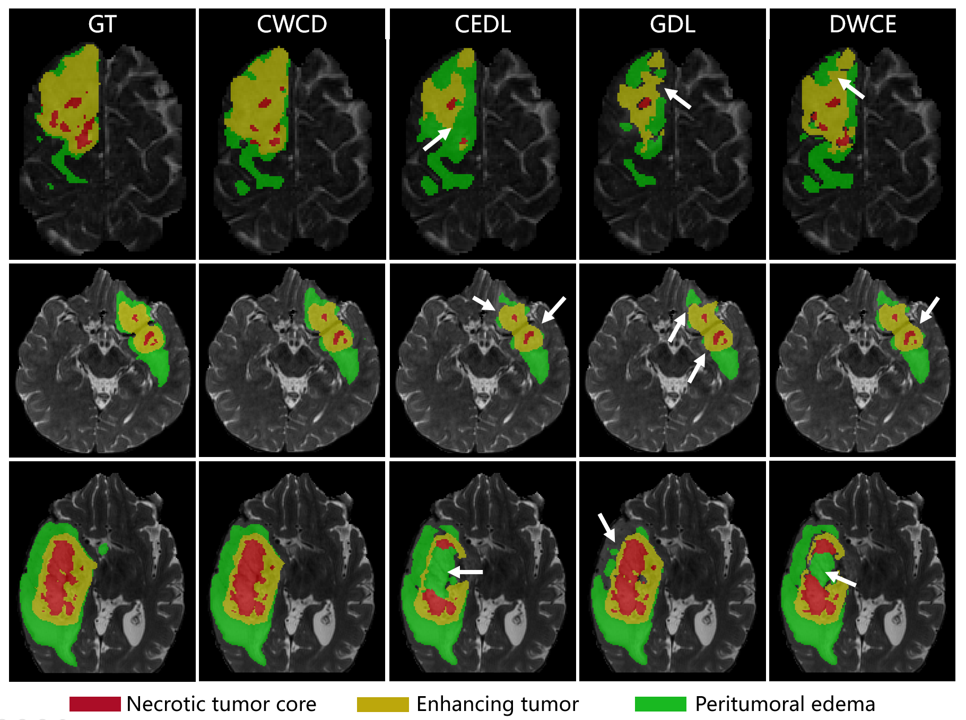}
    \caption{Qualitative comparison of different loss functions based on U-Net on the BraTS dataset. The whole tumor (WT) encompasses a union of red, yellow, and green regions. The tumor core (TC) includes the union of red and yellow regions. The enhancing tumor core (ET) denotes the green region.}
    \label{brats visual}
\end{figure}
\begin{figure*}[htbp]
    \centering
    \includegraphics[width=6.4 in]{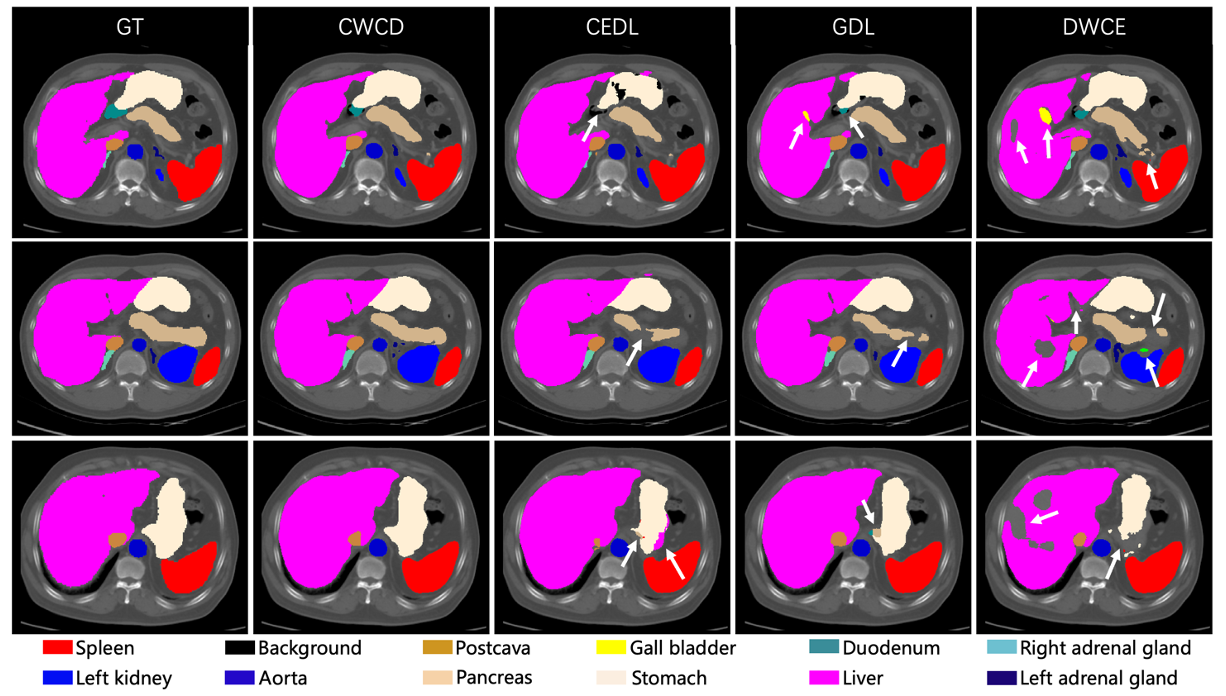}
    \caption{Qualitative comparison of different loss functions based on U-Net on representative slices from the AMOS dataset.}
    \label{abdominal visual}
\end{figure*}

\vspace{-0.2cm}
\section{RESULTS}
\label{Results}
\subsection{Assessments of Different Loss Functions}
To assess the performance of our proposed compound loss function (CWCD), we compared it with three other related loss functions denoted by GDL, DWCE, and CEDL, respectively.

We leveraged different loss functions to train the U-Net independently on each dataset with a consistent training protocol. The performance of these loss functions in terms of DSC on both datasets is depicted in Table \ref{Tab 1}. The results show that our proposed loss function offered the best performance on both datasets in terms of DSC, particularly on the AMOS dataset where the class imbalance issue was much more prominent. Our method outperformed the second-best (CEDL) by 3.94\%. Although our method (CWCD) marginally outperformed GDL on the BraTS dataset (0.8099 vs 0.8070), the performance gap on the AMOS dataset was considerably big (0.7497 vs 0.7046). Therefore, the results demonstrate that our proposed loss function performed better in the presence of severe data imbalance (on the AMOS dataset).

Fig. \ref{brats visual} shows the qualitative evaluations in three representative slices from the BraTS dataset. The results show that our proposed loss function achieved superior performance, as compared to its three peers. The segmentation results shown in the top and bottom rows convey the remarkable improvement in the accuracy by using our loss function. In the middle row, it shows that the three competing loss functions suffered from false positives while our loss function successfully identified all the target regions. Fig.\ref{abdominal visual} shows the qualitative assessments in abdominal organ segmentation. Overall, the results show that our proposed loss function (CWCD) yielded the most accurate segmentation, as compared to its three competing correspondences, in particular in the regions of pancreas, stomach, liver, and gall bladder.

\vspace{-0.2cm}
\subsection{Assessments with Different Networks}
We assessed our proposed loss function by injecting it into different networks, in order to evaluate its robustness across deep architectures. We adopted four semantic segmentation models: U-Net, V-Net, DeepLabV3 \cite{chen2017rethinking}, and UNETR \cite{hatamizadeh2022unetr}. All those models were not pre-trained. U-Net is the backbone network for medical image segmentation. V-Net is based on the U-Net framework, but with a residual module added. DeepLabV3 leverages multi-scale feature information through an atrous spatial pyramid pooling module. UNETR combines the advantages of CNN and Transformers \cite{vaswani2017attention}. 
\begin{table*}[htbp]
    \scriptsize
    \centering
    \caption{Segmentation results obtained by using the CEDL loss and ours (CWCD) with different networks on the two datasets in terms of DSC. Hyphen (-) indicates a complete failure in prediction.}
    \label{Tab 2}
    \renewcommand\arraystretch{1.05}
    \begin{tabular}{cccccccccc}
        \toprule
        \specialrule{0em}{0.2pt}{0.2pt}
        \cline{1-10}
        
        \multirow{2}{1cm}{Dataset}    & \multirow{2}{1cm}{Organ}
        &\multicolumn{2}{c}{U-Net\cite{ronneberger2015u}} & \multicolumn{2}{c}{DeepLabV3\cite{chen2017rethinking}}   &\multicolumn{2}{c}{UNETR\cite{hatamizadeh2022unetr}} & \multicolumn{2}{c}{VNet\cite{milletari2016v}} \\
        \cline{3-10}
        & & CEDL & CWCD & CEDL & CWCD & CEDL & CWCD & CEDL & CWCD \\ 
        \cline{1-10}
        & Sp &0.8831	&0.9023	&0.9330	&0.9425	&0.8575	&0.8710	&0.3958	&0.8218 \\
        & RK &0.8961	&0.9229	&0.9241	&0.9357	&0.8931	&0.8963	&0.7108	&0.8608 \\
        & LK &0.8875	&0.9198	&0.9399	&0.9382	&0.8926	&0.9036	&0.3897	&0.8701 \\
        & Ga &0.6416	&0.6248	&0.7134	&0.7324	&0.5176	&0.5253	&0.3241	&0.549  \\ 
        & Es &0.5170	&0.6290	&0.6386	&0.7434	&0.3986	&0.5467	&0.1784	&0.5071 \\ 
        & Li &0.9420	&0.9380	&0.9678	&0.9686	&0.9443	&0.9434	&0.9051	&0.9184 \\
        & St &0.7715	&0.8112	&0.9018	&0.8909	&0.6276	&0.5912	&0.8071	&0.8275 \\
AMOS    & Ao &0.9051	&0.9140	&0.9341	&0.9262	&0.8434	&0.8620	&0.7934	&0.8433 \\
        & Po &0.8288	&0.8374	&0.8658	&0.8680	&0.7372	&0.7703	&0.1752	&0.6854 \\
        & Pa &0.5974	&0.6814	&0.6628	&0.7458	&0.5072	&0.5656	&0.6558	&0.6712 \\
        & RAG&0.4397	&0.6067	&0.5230	&0.5817	&0.002	&0.5087	&-	&0.4747  \\
        & LAG&0.3549	&0.4844	&0.4661	&0.5818	&-	    &0.3276	&-	&0.3204  \\
        & Du &0.5942	&0.6372	&0.7326	&0.7562	&0.4233	&0.4830	&0.5881	&0.4977 \\
        & Bl &0.7120	&0.6844	&0.8215	&0.8247	&0.5939	&0.5621	&0.6279	&0.668  \\
        & P/U &0.6842	&0.6255	&0.7674	&0.7810	&0.3235	&0.1591	&0.2953	&0.5684 \\
        & Avg &0.7103	&0.7479(\textbf{+3.76\%})	&0.7861	&0.8145(\textbf{+2.84\%})	&0.6116	&0.6344(\textbf{+2.28\%})	&0.5267	&0.6723(\textbf{+14.56\%}) \\

        \cline{1-10}
        \rule{0pt}{8pt} 
        &WT	&0.8411	&0.8493	&0.8471	&0.8604	&0.7781	&0.7986	&0.7416	&0.8035  \\
BraTS   &TC	&0.7868	&0.8182	&0.8273	&0.8295	&0.7093	&0.7200	&0.6428	&0.7111  \\
        &ET	&0.7447	&0.7622	&0.7645	&0.7654	&0.6943	&0.7163	&0.6431	&0.6862  \\
        &Avg&0.7909	&0.8099(\textbf{+1.90\%})	&0.8130	&0.8184(\textbf{+0.54\%})	&0.7272	&0.7450(\textbf{+1.78\%})	&0.6758	&0.7336(\textbf{+5.78\%})\\     
\cline{1-10}
\specialrule{0em}{0.3pt}{0.3pt}
\bottomrule
    \end{tabular}
\end{table*}

We picked the CEDL loss as a baseline loss function, as it, similar to our proposed loss function, incorporates both cross-entropy loss and dice loss as well. We trained each model using CEDL and CWCD, respectively, and the results are shown in Table \ref{Tab 2}.

\vspace{-0.05cm}
The results show that our approach outperformed the baseline method on both datasets, in particular on the AMOS dataset. Our method considerably improved all the four models by 3.76\%, 2.84\%, 2.28\%, and 14.56\%, respectively, as compared to using the CEDL loss. Moreover, our approach substantially enhanced RAG and LAG predictions by over 10\% with U-Net and DeepLabV3. In contrast, CEDL predicted RGA and LGA on V-Net and UNETR models less accurately. If a model predicted a DSC of 0, \ie, a complete prediction failure, it was marked by a hyphen (-). To avoid the adverse impact of such predictions on the mean, we excluded the results with a predicted DCS of 0 when calculating the mean.

\vspace{-0.2cm}
\subsection{Ablation Study}
In order to comprehensively assess the contribution of each component in our proposed loss function, we conducted an ablation study. To ensure a fair and rigorous comparison, we used the U-Net architecture with identical parameters but underwent training with different components of our proposed loss function. We examined the components of cross-entropy (CE), contour-weighted cross-entropy (CWCE), dice loss (DL), and separable dice loss (SDL). This systematic approach allowed us to dissect the influence of each loss component on the model's performance.

Table \ref{Tab 3} shows the results of our ablation study on the two datasets. The results show that 1) the adoption of both our compound loss function and contour weighting map led to the enhancement in segmentation accuracy; 2) the cross-entropy loss performed as a more important component than the dice loss, underlining its substantial contribution to improved segmentation outcomes; and 3) despite a decrease in the performance of the dice loss function with an increase in the number of segmentation targets, its significance remained undiminished within the overarching framework of our proposed compound loss function.
\begin{table}[h]
    \scriptsize
    \centering
    \setlength{\abovedisplayskip}{-0.8cm} 
    \caption{Ablation study results of our proposed approach. The best performance per dataset in terms of DSC is highlighted in bold. Hyphen (-) indicates a complete failure in prediction.}
    \label{Tab 3}
    \begin{tabular}{ccccccc}
        \toprule
        \specialrule{0em}{0.2pt}{0.2pt}
        \cline{1-7}
        Dataset & Organ & CE & CWCE & DL & SDL & CWCD\\ 
        \cline{1-7}
        	& Sp &0.8679	&0.9069	&-	&-	&0.9023 \\
                & RK &0.9193	&0.9321	&-	&0.8988	&0.9229 \\
                & LK &0.8852	&0.9029	&-	&0.9151	&0.9198 \\
                & Ga &0.5722	&0.6577	&-	&-	&0.6248 \\ 
                & Es &0.4976	&0.6063	&-	&-	&0.6290 \\ 
                & Li &0.9350	&0.9464	&0.0936	&0.9399	&0.9380 \\
                & St &0.8194	&0.8021	&-	&0.7956	&0.8112  \\
        AMOS    & Ao &0.9035	&0.9134	&-	&0.8850	&0.9140  \\
                & Po &0.8263	&0.8146	&-	&0.8283	&0.8374  \\
                & Pa &0.6491	&0.6836	&-	&-	&0.6814  \\
                & RAG&0.5083	&0.5678	&0.5376	&-	&0.6067  \\
                & LAG&0.3566	&0.5611	&-	&-	&0.4844  \\
                & Du &0.6021	&0.6583	&-	&-	&0.6372  \\
                & Bl &0.6161	&0.6236	&-	&0.7073	&0.6844  \\
                & P/U &0.4097	&0.4435	&-	&-	&0.6255  \\
                & Avg &0.6912	&0.7347	&0.3156	&\underline{0.8529}	&\textbf{0.7479} \\

        \cline{1-7}
        \rule{0pt}{8pt} 
                &WT	&0.8187	&0.8435	&0.7615	&0.7954	&0.8493  \\
        BraTS   &TC	&0.7633	&0.7823	&0.6417	&0.6649	&0.8182  \\
                &ET	&0.7327	&0.7301	&0.7259	&0.7470	&0.7622  \\
                &Avg&0.7716	&0.7853	&0.7097	&0.7358	&\textbf{0.8099}  \\     
        \cline{1-7}
        \specialrule{0em}{0.3pt}{0.3pt}
        \bottomrule
    \end{tabular}
\end{table}

Although the inter-class and intra-class imbalance issues are more prominent on the AMOS dataset, our proposed approach yielded good segmentation results. As the dice loss (DL) is a region-based loss that focuses only on the target regions, its effectiveness decreased as the number of segmentation classes increased, as shown in Table \ref{Tab 3}, indicating that employing DL alone led to the worst result. In particular, on the AMOS dataset with more categories, using DL alone failed to predict most organs. Fortunately, when the loss function was switched to our proposed separable dice loss, the predictions were substantially improved. Although there were still some organs that could not be predicted, the mean DSC for the seven predicted organs reached 0.8529 (underlined). Therefore, both CWCE and SDL based on contour weighting maps generated considerably improved segmentation results.

\section{DISCUSSION}
\label{Discussion}
We have developed a novel compound loss function utilizing a contour weighting strategy based on dice loss and cross-entropy loss. We targeted addressing both intra- and inter-class imbalances in the challenging task of multi-object segmentation for medical images, and have assessed our approach by performing image segmentation for abdominal organs and brain tumors on two public datasets, respectively. The experimental results have demonstrated that our proposed approach offered substantially improved segmentation results in the two segmentation tasks, when collaborating with different deep networks, as compared to several other related loss functions.

Nevertheless, it is important to note that our approach in the current version employed uniform parameter settings for extracting contours of segmented organs and tumors, neglecting potential variations in shape and size effects. In addition, the contour extraction process was conducted offline, which limited the flexibility of our approach. In our future research endeavors, we would aim to delve deeper into the method and refine the contour extraction technique. Our focus would extend to exploring adaptive contour extraction training methods, with the ultimate goal of further improving segmentation performance. This ongoing work seeks to additionally promote the flexibility and robustness of our approach for medical image segmentation.



\bibliographystyle{IEEEbib}
\bibliography{strings,refs}

\end{document}